\documentclass{article}
\usepackage[utf8]{inputenc}
\usepackage[dvipsnames]{xcolor}
\usepackage{pdflscape}
\usepackage{color, colortbl}
\usepackage[toc,page]{appendix}
\usepackage[hang]{footmisc}
\usepackage{graphicx, fancyhdr, amssymb, amsmath, geometry, lscape, theorem, xcolor,float,rotating,setspace, authblk, longtable, ulem, enumitem}

\definecolor{Grayy}{gray}{0.95}
\usepackage{overcite,hyperref}
\usepackage{soul}
\usepackage{float}
\usepackage[dvipsnames]{xcolor}
\usepackage{fancyvrb}
\usepackage{etoolbox}
\usepackage{booktabs}
\usepackage{multirow}
\usepackage{soul}
\usepackage{ulem}

\usepackage{array,booktabs}
\newcolumntype{L}{@{}>{\kern\tabcolsep}l<{\kern\tabcolsep}}
\usepackage{colortbl}
\usepackage{xcolor}

\makeatletter %% <- make @ usable in command sequences
\newcount\SOUL@minus
\makeatother  %% <- revert @

\begin{document}
\onehalfspacing
\title{Improving Nonalcoholic Fatty Liver Disease Classification Performance With Latent Diffusion Models}

\author[1,*]{Romain Hardy}
\author[1]{Joe Klepich}
\author[1]{Ryan Mitchell}
\author[1]{Steve Hall}
\author[1]{Jericho Villareal}
\author[1,$\dagger$]{Cornelia Ilin}

\affil[1]{\small School of Information, U.C. Berkeley \vspace{1cm}}

\affil[*]{First author}
\affil[$\dagger$]{Corresponding author: cornelia.ilin@berkeley.edu}

\affil{\small School of Information, U.C. Berkeley}

\date{\today}

\maketitle

\begin{abstract}
\noindent
Integrating deep learning with clinical expertise holds great potential for addressing healthcare challenges and empowering medical professionals with improved diagnostic tools. However, the need for annotated medical images is often an obstacle to leveraging the full power of machine learning models. Our research demonstrates that by combining synthetic images, generated using diffusion models, with real images, we can enhance nonalcoholic fatty liver disease (NAFLD) classification performance even in low-data regime settings. We evaluate the quality of the synthetic images by comparing two metrics: Inception Score (IS) and Fr\'{e}chet Inception Distance (FID), computed on diffusion- and generative adversarial network (GAN)-generated images. Our results show superior performance for the diffusion-generated images, with a maximum IS score of $1.90$ compared to $1.67$ for GANs, and a minimum FID score of $69.45$ compared to $100.05$ for GANs. Utilizing a partially frozen CNN backbone (EfficientNet v1), our synthetic augmentation method achieves a maximum image-level ROC AUC of $0.904$ on a NAFLD prediction task.

\end{abstract}

\newpage
\doublespacing

\section*{\color{Maroon}Introduction}

\begin{figure}[H]
	\centering
	\includegraphics[width=\linewidth]{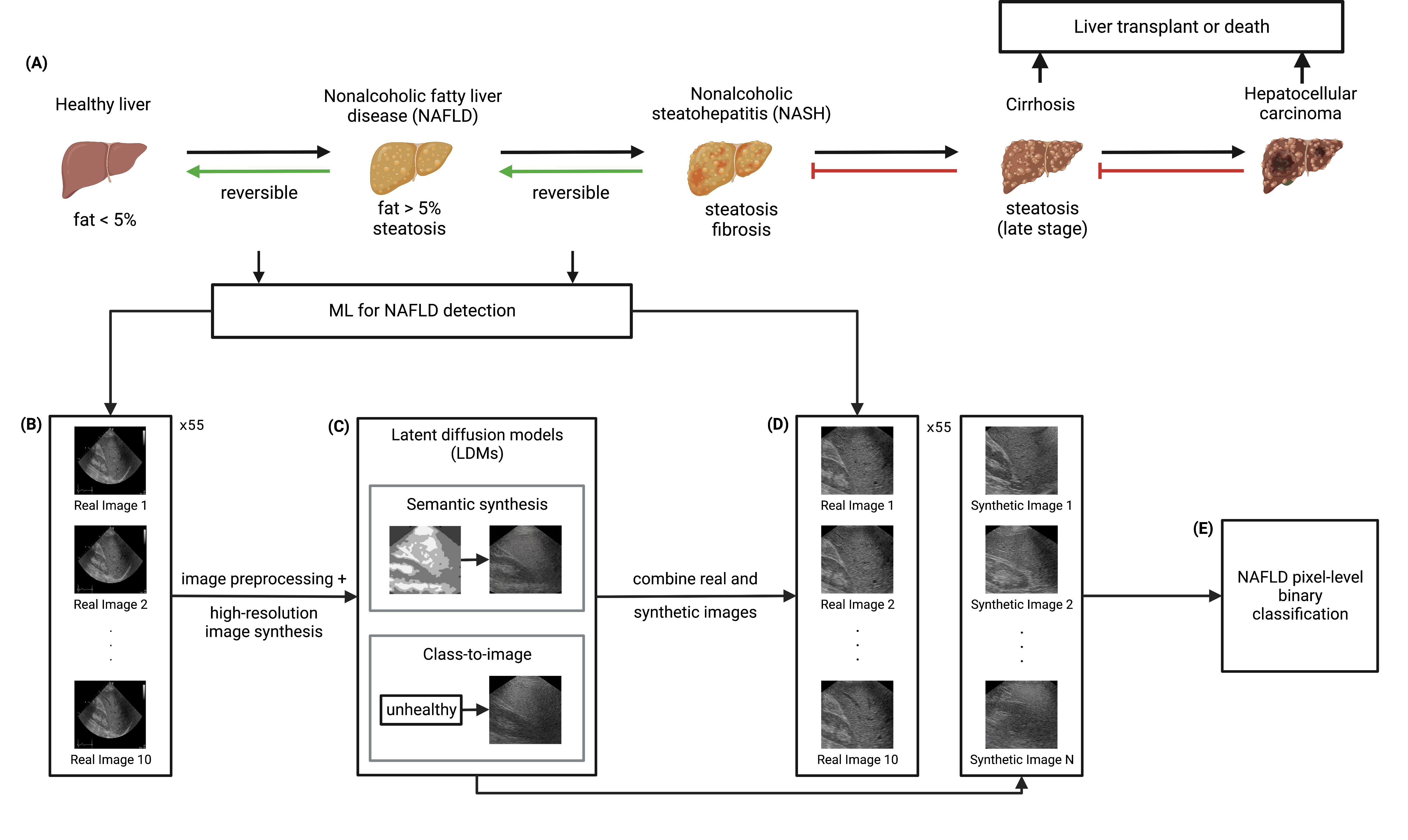}
	\caption{\label{fig:figure1}\textbf{An overview of NAFLD progression and an approach to combine liver ultrasounds and latent diffusion models for NAFLD detection.} \textbf{(A)} The disease progression from a healthy liver to NAFLD, NASH, cirrhosis, and ultimately HCC. The first two transitions are reversible through treatment and lifestyle changes (left-pointing green arrows), whereas the latter two can only be remedied using a liver transplant (left-ended red lines). \textbf{(B)} Examples of liver ultrasound images used in our analysis. The raw images are preprocessed before being fed to our downstream models (C-E). \textbf{(C)} We train two classes of latent diffusion models for liver ultrasound synthesis: semantic synthesis models and class-to-image models. Semantic synthesis models are conditioned on low-resolution semantic maps, while class-to-image models are conditioned on patient class labels. \textbf{(D-E)} For the NAFLD classification stage of our pipeline, we feed a mixture of real and synthetic ultrasound images to pretrained CNNs.}
\end{figure}

Nonalcoholic fatty liver disease (NAFLD) is a condition unrelated to alcohol consumption, whereby excess fat builds up in the liver without a clear cause. NAFLD progresses to nonalcoholic steatohepatitis (NASH), followed by cirrhosis and hepatocellular carcinoma (HCC). NAFLD and NASH can be reversed with adequate treatment and lifestyle changes. However, cirrhosis and HCC generally necessitate a liver transplant and can otherwise lead to death (see Figure \ref{fig:figure1}A). Understanding the incidence rate of HCC in patients at different stages of NAFLD progression is thus essential. Orci et al., 2021\cite{orci2022incidence} systematically review existing literature and find that for patients in the early stages of NAFLD, before cirrhosis, the incidence rate of HCC is 0.03 [95 percent CI: 0.01-0.07] per 100 person-years; in contrast, patients with cirrhosis have a significantly higher incidence rate of 3.78 [95 percent CI: 2.47–5.78] per 100 person-years; and the incidence rate for cirrhosis patients regularly screened for HCC is even higher at 4.62 [95 percent CI: 2.77–7.72] per 100 person-years. Given these statistics and the irreversibility of late-stage NAFLD, early diagnosis is a key step toward prevention of HCC. 

Liver biopsy is generally an effective method for detecting NAFLD, but it involves a procedure that is both invasive and expensive (e.g., Gaidos et al., 2008\cite{gaidos2008biopsy}, Villani et al., 2023\cite{villani2023nafld}). By comparison, ultrasound imaging is a fast, safe, and cheap method. Its disadvantage lies in its low specificity relative to liver biopsy, especially since the quality of ultrasound images is variable and depends on the ability of the ultrasound technician (e.g., Strauss et al., 2007\cite{strauss2007nafld}, Khov et al., 2014\cite{khov2014nafld}, and Acharya et al., 2016\cite{acharya2016nafld}). 

Recently, there has been a pronounced effort to use deep learning methods to improve the performance of diagnostic techniques for liver ultrasounds (e.g., Liu et al., 2017\cite{liu2017clf}, Biswas et al., 2017\cite{biswas2017dl}, Meng et al., 2017\cite{meng2017cnn}, Reddy et al., 2018\cite{reddy2018cnn}, and Che et al., 2021\cite{che2021cnn}). As with other medical imaging tasks, one of the key challenges hindering this effort is the lack of labeled data necessary to train large models. Collecting and annotating medical data is an expensive and difficult task that requires professional oversight. Even when a sizeable dataset is amassed, its use is often restricted by licenses and protocols to protect patient anonymity. An alternative is to synthesize realistic images, and a common approach for producing such data is through the use of geometric transformations such as rotations and flips (e.g., Garcea et al., 2013\cite{garcea2023}). However, this approach is limited in its ability to produce meaningful diversity and improve classification performance. More recently, research has focused on generative models to produce diverse and high-quality synthetic images from scratch. For example, Che et al., 2021\cite{che2021gan} show that generative adversarial networks (GANs) can be used to produce synthetic liver ultrasound images and improve classification accuracy of NAFLD. However, the diversity of the images produced by the authors is limited, and they do not compare their approach to more traditional geometric augmentation techniques. 

In this study, we utilize latent diffusion models (LDMs), cutting-edge generative models introduced by Rombach et al., 2021\cite{rombach2021ldm}, to create synthetic liver ultrasounds starting from a low-data regime setting. The LDM architecture not only showcases excellent performance in image generation tasks but also enables us to circumvent certain limitations associated with GANs. We adopt B-mode liver ultrasounds as our training data for two LDMs: the first LDM is conditioned on low-resolution semantic maps, and the second is conditioned on patient class labels (see Figure \ref{fig:figure1}B-C). Subsequently, we employ the trained LDMs to generate synthetic liver ultrasound images, which are then combined with real images. This union of real and synthetically generated images is then used for pixel-level binary classification of NAFLD using pretrained CNN backbones, specifically ResNet, EfficientNet v1, and EfficientNet v2, as shown in Figure \ref{fig:figure1}D-E.

This paper contributes to the existing literature as follows: first, our two latent diffusion models show better performance on the Inception Score (IS) and Fr\'{e}chet Inception Distance (FID) metrics than the GAN models proposed by Che et al., 2021\cite{che2021gan} and are able to reproduce key visual characteristics of NAFLD even in a data-constrained environment; second, we find that mixing real and synthetic ultrasound images improves the ROC AUC performance on held-out data for the NAFLD classification task; third, we demonstrate that the performance gain obtained from using synthetic images is consistent across the three CNN architectures mentioned above, and is larger in magnitude than that obtained by using traditional geometric augmentation techniques.

\section*{\color{Maroon}Data}
\label{data_section}
This study uses a dataset of B-mode liver ultrasounds acquired by the Department of Internal Medicine, Hypertension and Vascular Diseases at the Medical University of Warsaw, Poland (Byra et al., 2018 \cite{byra2018dataset}). The ultrasounds were collected from severely obese patients (mean age $40.1 \pm 9.1$, mean BMI $45.9 \pm 5.6$) admitted for bariatric surgery. The steatosis level was determined by pathologists based on the percentage of hepatocytes (obtained by wedge liver biopsy) with fatty infiltration; livers with more than 5\% infiltration were labeled fatty. Of the 55 patients, 17 were classified as healthy (non-fatty) and 38 were classified as unhealthy (fatty) (see Supplementary Table \ref{fig:ET0} in Appendix~\ref{data_analysis}). The ultrasound images were acquired using the GE Vivid E9 Ultrasound System equipped with a sector probe operating at 2.5 MHz. A sequence of 10 consecutive B-mode images of resolution $434 \times 636$ was collected for each patient, for a total of 550 images (see Figure \ref{fig:figure1}B for an example). Additional information about the data is presented in Byra et al., 2018\cite{byra2018dataset}.

\subsection*{Data Analysis}
In this section, we explore the presence of stylistic disease features within our real ultrasound images to assess their impact on the reconstruction (synthesis) of images related to NAFLD. To begin, healthcare professionals commonly employ several semi-quantitative scoring systems to assess and grade NAFLD, with the most prevalent ones being the Fatty Score (FS), the Ultrasonographic Fatty Liver Indicator (US-FLI), and the Hamaguchi Score (HS) (e.g., Ballestri et al., 2020 \cite{ballestri2020semi}). These scoring systems primarily evaluate specific stylistic characteristics associated with NAFLD, which encompass: (a) increased hepatic echogenicity (heightened contrast between the liver and the kidney due to the accumulation of lipids in the liver, causing the transducer beam to reflect and the liver to appear brighter, or hyperechoic, relative to the kidney), (b) blurred hepatic vein borders (fatty deposition can cause the outlines of hepatic veins to appear ill-defined in ultrasounds), and (c) blurred diaphragms (the boundary of the diaphragm, which separates the chest and abdominal cavities, may be ill-defined in patients with NAFLD). Dr. Joe Klepich, the medical expert on our team, conducted a thorough analysis of the real images in our dataset and confirmed the presence of these defining characteristics (see Supplementary Table \ref{fig:ET0} in Appendix~\ref{data_analysis}). Furthermore, if our actual ultrasound images happen to exhibit stylistic features associated with other diseases due to comorbidities linked to NAFLD or unrelated conditions, we anticipate improved generalization performance in the tasks of NAFLD image reconstruction and classification. Lastly, we recognize that analyzing large-scale datasets is usually the optimal approach for capturing diverse image- and patient-related features of NAFLD. However, unlike other research in medical image synthesis (e.g., Kim et al., 2022 \cite{kim2022retina}), our study is constrained by the limited availability of public data documenting NAFLD. Nonetheless, we will demonstrate later in the paper that, despite this limitation, our dataset still enables us to push the boundaries of state-of-the-art NAFLD image synthesis and classification through machine learning.

\subsection*{Data Preprocessing}
Considering the small number of images in our dataset, we opt for a 5-fold validation strategy when conducting experiments. Due to the high level of correlation between images from the same patient, we use a patient-level split to assign folds; thus, each fold has 44 patients (440 images) in the training set and 11 patients (110 images) in the validation set. We also stratify the folds by class label to maintain a consistent class distribution throughout (see Figure \ref{fig:figure2}A-B). Following the cross-validation split, we apply the data preprocessing logic outlined below before feeding the real ultrasound images to the two latent diffusion models and, ultimately, to the CNN models used for NAFLD classification (see Figure \ref{fig:figure2}C).

Concretely, we extract the region of interest from each raw image by cropping the largest connected component; this step removes text and instrumental annotations while keeping the ultrasound pixels. Then, we center crop the images to $320 \times 320$, maintaining the original pixel spacing of the images. Finally, we linearly scale the images to $[-1, 1]$ and apply a random resized crop to further reduce the image size to $256 \times 256$. The randomness introduced by this last step adds diversity at training time, which prevents our models from overfitting due to the relatively small sample of images. Appendix~\ref{data_preprocessing} contains additional data preprocessing details.

\section*{\color{Maroon}Methods}
\label{methods}

This study has two objectives. The first objective is to train two latent diffusion models to produce high-quality synthetic liver ultrasounds. The second objective is to show that mixing synthetic images produced by diffusion models with real data meaningfully improves performance on a NAFLD classification task. We describe our methodology below.

\subsection*{Latent Diffusion Models}

Recently, diffusion models --- a family of probabilistic models that learn to model distributions by recovering signals from noisy data (e.g., Ho et al., 2020\cite{ho2020pdm}) --- have demonstrated state-of-the-art performance across multiple image generation tasks. Compared to previous approaches such as generative adversarial networks (GANs), they are simpler to optimize and do not suffer from instabilities related to adversarial training. Our focus is on the recently introduced latent diffusion models (LDMs) of Rombach et al., 2021\cite{rombach2021ldm}, variants of diffusion models that first encode images to a low-dimensional latent space representation using a pretrained encoder, thus avoiding costly computations in pixel space. Appendix~\ref{latent_diffusion_models} contains details of the LDM modeling approach.

The LDMs we use contain autoencoders pretrained on the OpenImages dataset, with denoising UNet backbones that we train from scratch. Additionally, the LDM architecture allows us to condition the image generation process on a variety of inputs such as text, semantic maps, and class labels. The conditioning inputs are first projected to a latent representation space, and are then combined with intermediate layers of the UNet module via cross-attention or concatenation layers. We take advantage of this architectural design to fine-tune two LDMs, each conditioned on a different type of input. The first LDM is a semantic synthesis model conditioned on low-resolution semantic maps, while the second is a class-to-image model conditioned on patient class labels. Unlike previous approaches to medical image synthesis, we choose to employ multiple conditioning mechanisms, as this allows us to quantitatively assess their advantages and disadvantages, especially given our data-constrained environment.

\subsubsection*{\textcolor{Gray}{Synthetic Image Generation}}
\label{gen_synth_img}
The semantic synthesis LDM is conditioned on $128 \times 128 \times 5$ semantic maps, and uses a pretrained autoencoder with latent dimension $3 \times 64 \times 64$. The semantic maps are projected to this latent space and concatenated with the encoded image representations before being fed through the UNet module. To generate the semantic maps, we threshold the original images using Otsu's method (Otsu et al., 1979\cite{otsu1979}) and downsample the resulting masks by a factor of two. Ideally, the semantic maps should be created by medical experts with labels for important anatomical structures (such as the liver, kidney, diaphragm, portal veins, and hepatic veins) as separate classes. In the absence of professionally annotated images, however, our thresholding approach is a reasonable alternative. Each pixel in the ultrasound is assigned a class according to its relative brightness level, which approximately corresponds to the tissue density at that location in the human body. The class-to-image LDM is conditioned on the class label associated with each patient: $0$ for healthy (non-fatty) and $1$ for unhealthy (fatty). This model uses a pretrained autoencoder with a latent dimension $4 \times 32 \times 32$. The class label conditions are fed through an embedding layer of dimension $512$ before being meshed with the denoising UNet model via a cross-attention layer.

\begin{figure}[H]
	\centering
	\includegraphics[width=\linewidth]{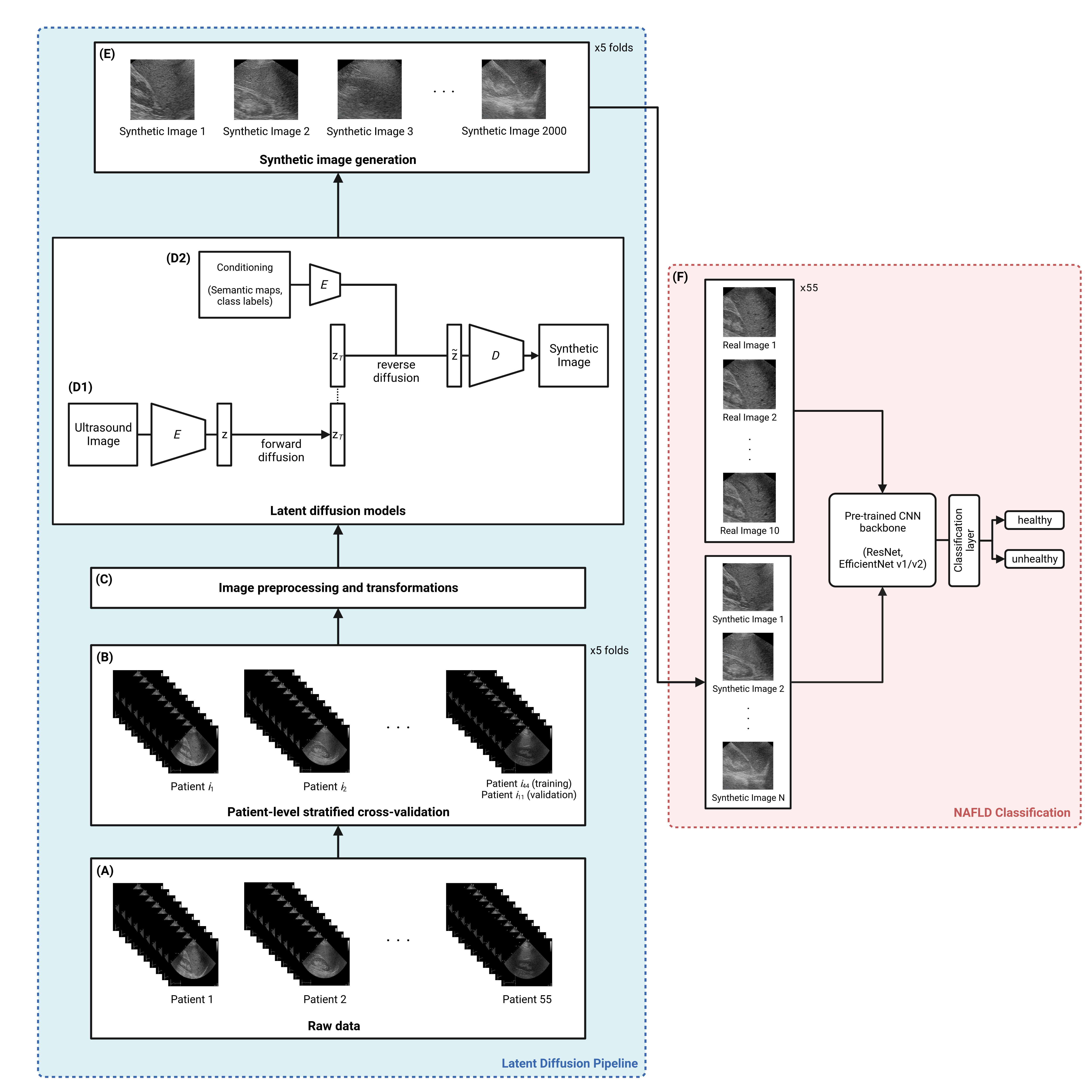}
	\caption{\label{fig:figure2}\textbf{An approach to use latent diffusion models to generate synthetic liver ultrasounds.} \textbf{(A-B)} The raw images are split into 5 folds at the patient level, stratified by class label. \textbf{(C)} The raw images are then preprocessed and fed to LDMs as training inputs. \textbf{(D.1-2)} For each fold, we train two diffusion models, a semantic synthesis LDM and a class-to-image LDM. During training, the model adds noise to the latent representations of input images and learns to recover the original signals using conditioning inputs as guidance. \textbf{(E)} After training, we generate new ultrasound images by randomizing the latent representations and conditioning inputs. For each diffusion model and each fold, we produce a sample of 2,000 synthetic images. \textbf{(F)} We mix synthetic and real ultrasounds to fine-tune pretrained backbones for a binary NAFLD classification task.}
\end{figure}

During training, our real (preprocessed) ultrasound images are embedded to a latent representation $z$ using a frozen pretrained encoder $E$ and corrupted by the forward diffusion process, producing $z_T$ (see Figure \ref{fig:figure2}D.1). Then, the denoising UNet model learns to reverse this process (guided by conditioning inputs, which are semantic maps for the semantic synthesis model and class labels for the class-to-image model), producing $\tilde{z}$, which is projected back to pixel space using a frozen decoder $D$ (see Figure \ref{fig:figure2}D.2). 

For a given fold, each LDM is trained for 20,000 steps with a batch size of $32$ on a Nvidia GeForce GTX 4090 GPU. We use the AdamW optimizer with a weight decay factor of $0.01$, and a learning rate of $2.4 \times 10^{-4}$ on a cosine schedule (100 warm-up steps). After training, we use the LDMs to produce samples of synthetic ultrasound images for each of the five folds (see Figure \ref{fig:figure2}E). Specifically, we randomize the initial latent representations and conditioning inputs, and feed these through the denoising UNet and decoder modules. For the semantic synthesis LDM, we use semantic maps from the training patients as the conditions, distorted using a weak piecewise affine transformation. For the class-to-image LDM, we use an evenly distributed mix of positive and negative class labels as the conditioning inputs. In both cases, we use a Denoising Diffusion Implicit Models (DDIM) sampler (Song et al., 2022\cite{song2022denoising}) with $500$ inference steps and a classifier-free guidance scale (Ho et al., 2022\cite{ho2022classifierfree}) of $1.2$ to generate images. In total, we produce 2,000 synthetic images per fold, for a total of 10,000 synthetic images. We use these synthetic images to assess the quality of our diffusion models, as well as for our downstream classification experiments (see Figure \ref{fig:figure2}F).

\subsubsection*{\textcolor{Gray}{Synthetic Image Evaluation}}
\label{eval_synth_img}

We assess the quality of the synthetic images produced by the two LDMs through both qualitative and quantitative approaches. In our qualitative evaluation, we assess visual quality and conduct an image Turing test involving a panel of five medical professionals, comprising four general practitioners and one epidemiologist. Concretely, we employ 20 real images randomly chosen from the original dataset alongside 30 synthetic images generated by our LDMs (15 from the semantic synthesis model and 15 from the class-to-image model). Among these images, 31 represent the unhealthy class (13 real, 18 synthetic), while 19 represent the healthy class (7 real, 12 synthetic). These images are then randomly displayed on a web page\cite{TuringTestWebsite} (see Supplementary Figure \ref{fig:EF1} for a screenshot), and participants are tasked to determine the authenticity of each image, i.e., whether it is real or synthetic. To maintain consistency, the order of image presentation remains the same for all participants. Additionally, we do not provide any information about the ratio of synthetic to real images, the disease labels associated with the images, or any supplementary annotations that might bias the participants' judgements. To assess the results of the Turing test, we present the average accuracy, specificity, and sensitivity for each image class subgroup (unhealthy and healthy) and for the overall sample, calculated across all test participants. In this context, sensitivity refers to correctness in identifying real images, while specificity refers to correctness in identifying synthetic images.

In our quantitative evaluation, we report the Inception Score (IS) and Fréchet Inception Distance (FID) metrics. The IS score is computed using the logit outputs of an Inception v3 model pretrained on ImageNet-1k\cite{heusel2017gans, InceptionV3TF}, where higher scores are given to label distributions that have low entropy and are uniformly distributed across all possible labels (Salimans et al., 2016\cite{salimans2016is}). The FID score is computed using the same Inception v3 model\cite{TorchFID}, but evaluates synthetic images against a sample of real ``ground truth'' images. Specifically, the FID score compares the distributions of the synthetic and real images after having been fed through a certain number of layers of Inception v3 (we use the last hidden layer of dimension 2048 for our evaluation); low FID scores signify a close match between the synthetic and real samples (Heusel et al., 2017\cite{heusel2017fid}). In our calculation of the FID score, the real samples comprise the validation images for a given fold.

\subsection*{Classification Models}
\label{clf_models}

We conduct a variety of experiments to assess whether synthetic liver ultrasound images generated by LDMs can meaningfully improve NAFLD classification performance. We use the same patient-level folds outlined in the Data section to train and evaluate different CNN classifiers. Notably, when training a classifier on fold $k$ we only use synthetic images generated by a diffusion model trained on $k$; this ensures that no information from the held-out patients unfairly leaks into the synthetic data. For all experiments, we define the mixing rate $r$ as the proportion of synthetic images to real images in the training set. For example, $r = 0$ signifies that the classifier has been trained on only real images, while $r = 1$ signifies that the classifier has been trained on an equal number of real and synthetic images. In each experiment, we fix $r$ and train a classifier for a set number of steps. Subsequently, we take a snapshot of the model after the final training iteration and use it to make predictions on out-of-fold images; we use these out-of-fold predictions to compute the corresponding ROC AUC. We repeat this process across 25 random seeds to ensure that our results are robust. Furthermore, we calculate SHapley Additive exPlanations (SHAP\cite{SHAP}) with the initial fold and seed of our preferred CNN classifier. This allows us to (a) investigate the synthetic image attributes that influence positive and negative predictions of NAFLD and (b) assess the effectiveness of our classifiers in capturing the stylistic features of NAFLD identified in the Data section.

\section*{\color{Maroon}Results}
\label{results}

We first present results evaluating the quality of the synthetic images produced by our two LDMs. We then demonstrate the effect of increasing the mixing rate $r$ on NAFLD classification performance for ResNet-50 (He et al., 2015\cite{he2015resnet}), EfficientNet v1 (Tan et al., 2020\cite{tan2020effnetv1}), and EfficientNet v2 (Tan et al., 2021\cite{tan2021effnetv2}) models. 

\subsection*{{Synthetic Image Quality}}
\label{qual_synth_img}

We qualitatively assess the quality of the synthesized images using visual inspection and the image Turing test we introduced in the previous section.

\begin{figure}[H]
	\centering
	\includegraphics[width=\linewidth]{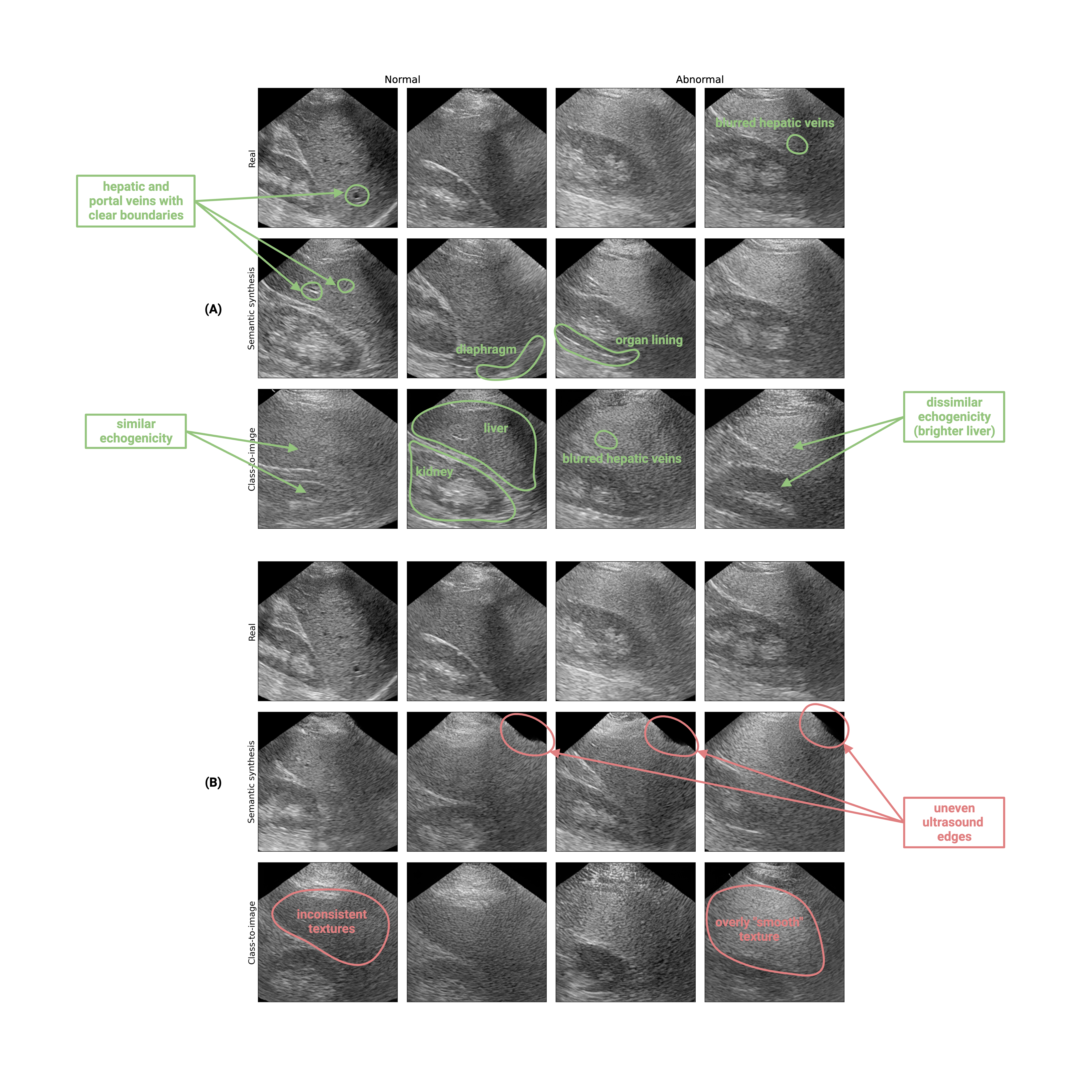}
	\caption{\label{fig:figure3}\textbf{Qualitative evaluation of semantic synthesis and class-to-image liver ultrasound images.} \textbf{(A)} Examples of high-quality synthetic images, with relevant anatomical structures and stylistic features annotated in green. \textbf{(B)} Examples of low-quality synthetic images, with aberrations annotated in red.}
\end{figure}

Visual illustrations of synthetic liver ultrasounds generated by the two families of LDMs (as described in Methods\ref{methods}) are presented in Figure \ref{fig:figure3}. Overall, we observe that both models can accurately reproduce major anatomical structures, such as the kidney and the liver. In some images, we also perceive finer details like hepatic veins, portal veins, and organ linings (see Figure \ref{fig:figure3}A). Most importantly, the diffusion models successfully replicate the essential stylistic features of NAFLD that we previously examined in the Data section. In synthetic examples belonging to the unhealthy class, we notice pronounced contrast levels between the liver and the kidney cortex, as well as blurred hepatic veins. Conversely, in healthy synthetic samples, the echogenicities of the liver and kidney are generally consistent and the vein boundaries are better defined. While we do observe the presence of the diaphragm in some synthetic images, it is challenging to ascertain whether the LDMs can accurately replicate the expected blurriness of this organ in unhealthy patients. However, given the limited number of training examples featuring visible diaphragms (see Supplementary Table \ref{fig:ET0} in Appendix \ref{data_analysis}), we believe that this limitation is understandable. Visual inspection also reveals that certain images have obviously unrealistic features. For instance, the texture of some samples produced by the class-to-image LDM appears either inconsistent or too ``smooth" compared that of to real ultrasound images. Other samples generated by the semantic synthesis LDM have uneven edges, which immediately expose them as fake (see Figure \ref{fig:figure3}B). This may be due to aberrant conditioning inputs rather than a faulty model, however. Strong piecewise affine transformations may adversely distort parts of the semantic map, causing the diffusion model to produce unrealistic images. 

The Turing test results, representing the assessment of image realism by five medical professionals, are summarized in Table \ref{fig:table1}. Across all 50 images presented, the participants achieved an average accuracy, specificity, and sensitivity of $51.2\%$ [$95$ percent CI: $40.6–61.8\%$], $59.0\%$ [$95$ percent CI: $32.2–85.8\%$], and $46.0\%$ [$95$ percent CI: $20.9–71.1\%$], respectively. These findings indicate that the participants were unable to consistently differentiate between real and synthetic images. Additionally, these results are consistent across both image class subgroups (unhealthy and healthy). Supplemental Figure \ref{fig:EF2} displays images that were frequently identified correctly and incorrectly by the participants.

\begin{table}[H]
\centering
\begin{tabular}{@{}l  c  c  c @{}}
\hline
 \textbf{Class}     & \textbf{Accuracy [$95\%$ CI]}   & \textbf{Sensitivity [$95\%$ CI]} & \textbf{Specificity [$95\%$ CI]} \\[0.1cm]
\hline
unhealthy & 0.510 [0.332, 0.687] & 0.569 [0.339, 0.799]  & 0.467 [0.164, 0.770]  \\[0.2cm]
healthy   & 0.516 [0.461, 0.570] & 0.629 [0.240, 1.017]  & 0.450 [0.170, 0.730]  \\[0.2cm]
both classes & 0.512 [0.406, 0.618] & 0.590 [0.322, 0.858]  & 0.460 [0.209, 0.711]\\
\hline
\end{tabular}
\caption{\label{fig:table1}\textbf{Turing test results.} We select a sample consisting of 20 real images, 15 semantic synthesis LDM images, and 15 class-to-image LDM images. These images are then assigned for realism assessment to five medical professionals. The table shows the average accuracy, sensitivity, and specificity across participants, for each disease class separately and for the full sample. Numbers in brackets represent $95\%$ confidence intervals (CI).}
\end{table}

\begin{figure}[H]
	\centering
	\includegraphics[width=\linewidth]{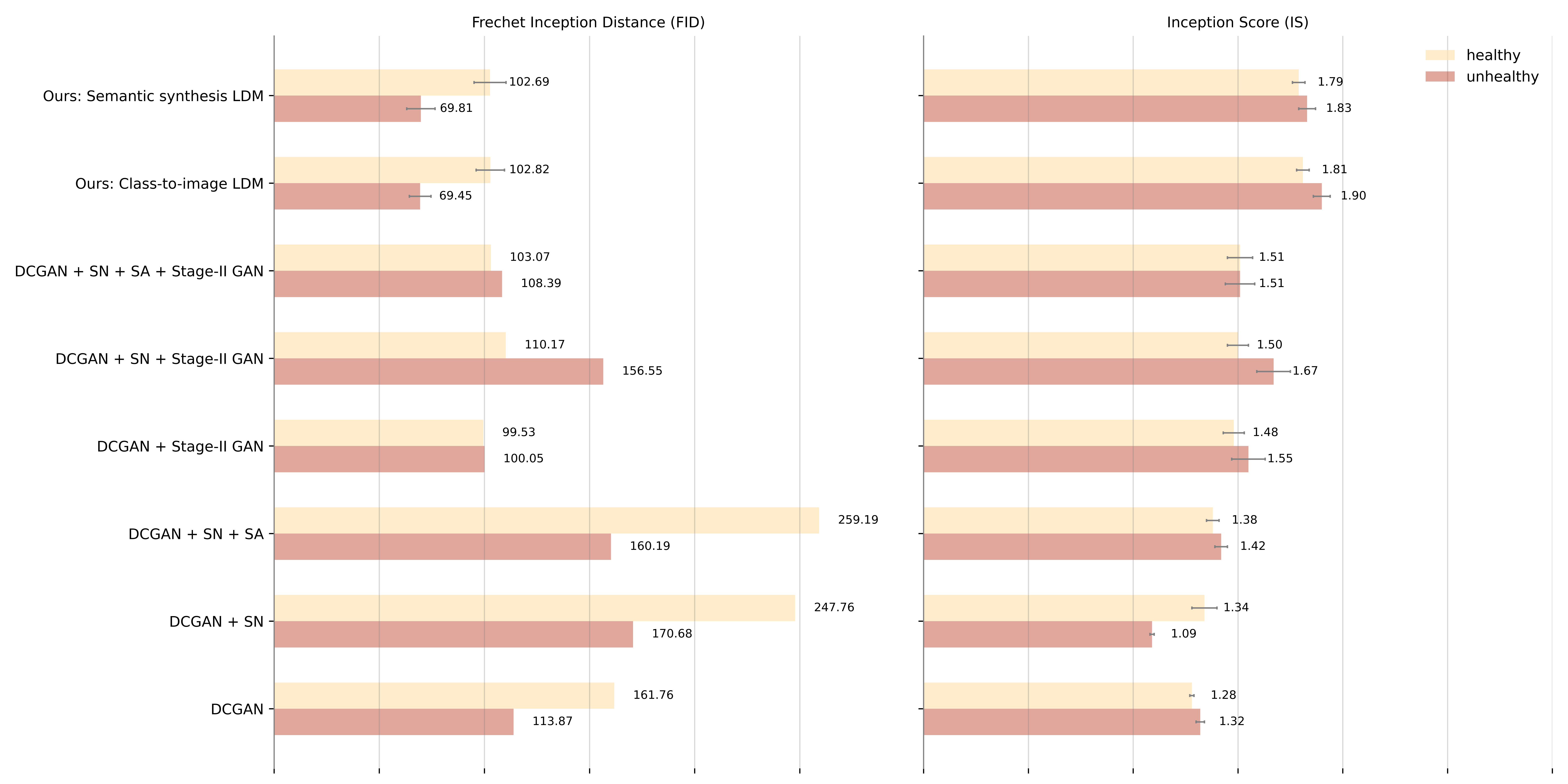}
	\caption{\label{fig:figure4}
 \textbf{Quantitative evaluation of semantic synthesis and class-to-image liver ultrasound images.} We report the FID and IS scores for each LDM and for each NAFLD class (horizontal bar plots 1-2) and compare our results against GAN architectures as reported by Che et al., 2021\cite{che2021gan} (horizontal bar plots 3-8). The horizontal gray lines added to the bar plots represent 95\% confidence intervals; no error estimates are provided for the GAN FID scores. \textbf{Note}: The comparison of our work to Che et al., 2021\cite{che2021gan} is not exactly one-to-one. Our results are averaged across five folds, while theirs are only reported over a single split. Also, the authors oversample healthy ultrasounds to create an equal balance of classes, whereas we leave the original class distribution untouched. Despite these differences, we choose to include their results to draw a comparison with our models.}
\end{figure}

For a quantitative evaluation of our synthetic images we report the IS and FID scores computed for each of the two LDMs, using samples of 10,000 synthetic images with 2,000 images per fold. We also include the scores of the GAN models as introduced by Che et al., 2021\cite{che2021gan} for an approximate comparison (see Figure \ref{fig:figure4}). We find that LDM-generated images generally outperform GAN-generated images across both metrics, demonstrating improved quality. 
We observe that our LDMs have superior performance on unhealthy ultrasounds, especially with regard to the FID score. This is likely due to the fact that our dataset contains more unique images belonging to unhealthy patients. Thus, since the LDMs are trained on a wider variety of abnormal liver ultrasounds, they are better at modeling this class' distribution. Although there is a difference between the disease classes, the FID and IS scores of both LDMs are remarkably similar. One possible interpretation is that the characteristics of our conditioning inputs play a secondary role in the context of NAFLD image reconstruction. However, it may also suggest that these metrics are inadequate for effectively distinguishing between the two LDMs, especially in settings with limited data. Later on, we rely on the outcomes of a NAFLD classification task as an additional means of comparing the conditioning mechanisms. Finally, we also attempt to balance the two classes by oversampling the minority (healthy) class to get more insights into model performance. We find that the IS and FID results are quantitatively similar with our original experiments (IS: \{Semantic synthesis LDM: 1.79 healthy, 1.81 unhealthy; Class-to-image: 1.84 healthy, 1.89 unhealthy\}, FID: \{Semantic synthesis LDM: 104.83 healthy, 72.95 unhealthy; Class-to-image LDM: 104.29 healthy, 69.77 unhealthy\}).

\subsection*{NAFLD Classification}
Our initial architecture consists of a ResNet-50 classifier pretrained on the ImageNet-1k dataset and fine-tuned for a NAFLD classification task. We conside four different NAFLD data input scenarios: one model is fine-tuned on real images only (base), and the other three are fine-tuned by mixing synthetic images and real images according to a nonzero mixing rate $r$. For the latter three scenarios, we use a mix of semantic synthesis and real images, a mix of class-to-image and real images, and a mix of geometrically-augmented and real images (traditional). The geometrically-augmented images are random resized crops of the original images, rotated between $-5^\circ$and $5^\circ$ with a $25\%$ chance. 

\begin{figure}[H]
	\centering
	\includegraphics[scale=0.35]{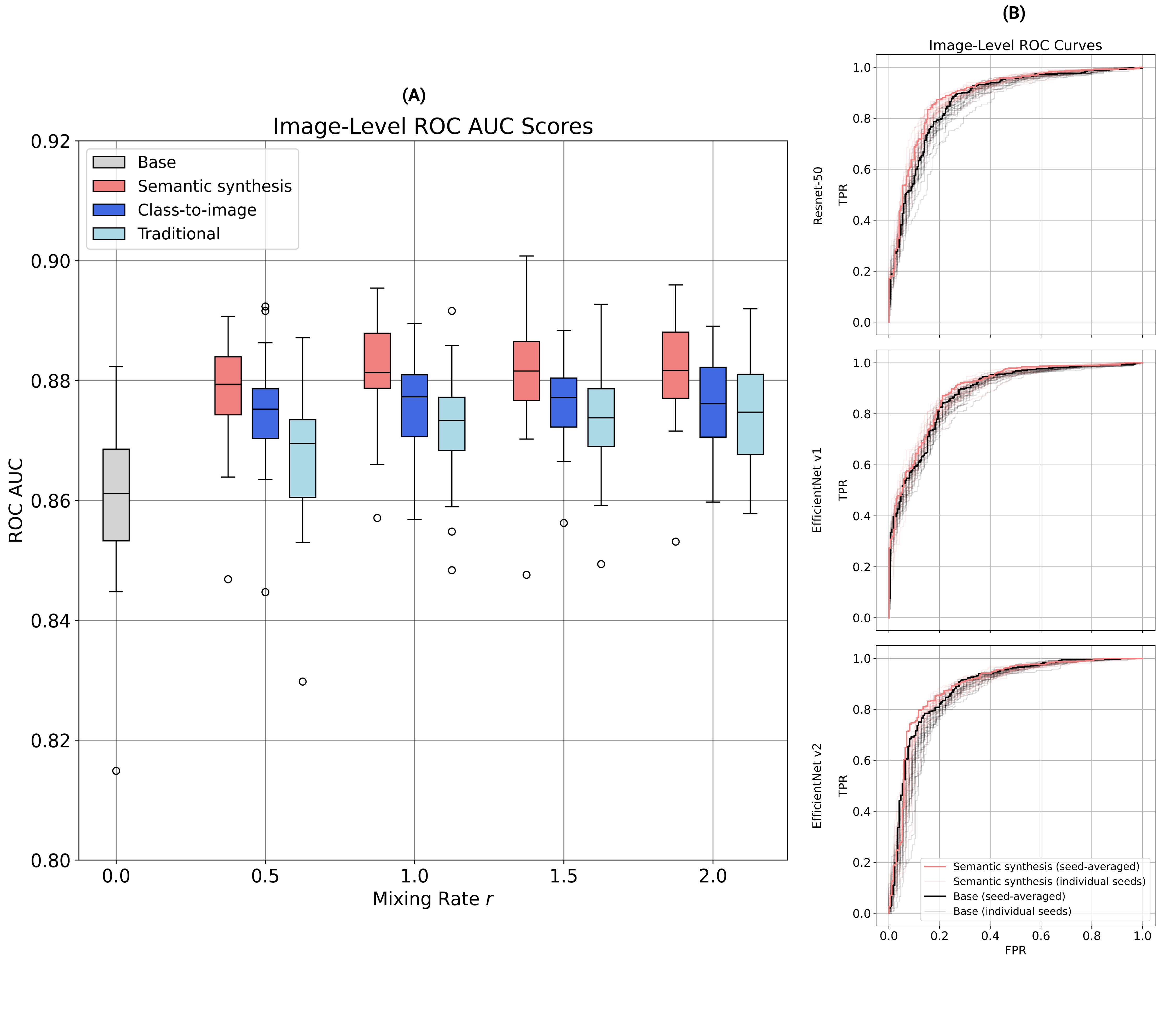}
	\caption{\label{fig:figure5}\textbf{Out-of-sample NAFLD classification performance as a function of the mixing rate $r$.} \textbf{(A)} Box plots of the image-level ROC AUC as a function of the mixing rate $r$, where $r$ ranges from 0 to 2 in intervals of 0.5. Each color corresponds to a different type of image data used for training: real images only (grey), a mix of semantic synthesis and real images (red), a mix of class-to-image and real images (dark blue), and a mix of geometrically-augmented and real images (light blue). The distribution over ROC AUC values for each type of input data is due to the repetition of each experiment with different random seeds. \textbf{(B)} Image-level ROC curves for three families of CNN classifiers: ResNet-50 (top), EfficientNet v1 (middle), and EfficientNet v2 (bottom). The black lines show the performance of each model when trained on real images only ($r = 0$), while the red lines show the model performance when trained on a mix of semantic synthesis and real images ($r = 1.5$). The bold lines are seed-averaged ROC curves, while the light lines correspond to individual seeds. The average ROC AUC for the base and synthetically augmented models are as follows: ResNet-50, 0.861 versus 0.882; EfficientNet v1, 0.871 versus 0.884; EfficientNet v2, 0.864 versus 0.879.}
 \end{figure}

For each input scenario, we freeze the first three convolutional blocks of ResNet-50 and fine-tune the weights corresponding to the neck, the fourth block, and the final classification layer. To isolate the effect of the mixing rate $r$ on model performance, the unfrozen layers are fine-tuned using the same hyperparameters (1,000 training steps with a learning rate of $2.0 \times 10^{-5}$ and a batch size of $32$) for each value of $r$. In Figure \ref{fig:figure5}A we report the image-level ROC AUC for each of the four data input scenarios, computed by concatenating the out-of-fold predictions for all five folds and then calculating the score on the combined predictions. To evaluate the statistical significance of our results, we repeat each experiment with $25$ different random seeds. 

Our results indicate that mixing synthetic images with real images steadily improves the classification performance of our fine-tuned ResNet-50 model, albeit with diminishing returns. We also find that the model trained on a mix of semantic synthesis and real images outperforms the other models. When compared to the base model, the greatest increase in classification performance for the semantic synthesis mix occurs at $r=1.5$ (average ROC AUC = $0.861$ versus average ROC AUC = $0.882$). We confirm that this result is statistically significant by conducting a paired $t$-test using ROC AUC scores from the same random seed as pairs. We observe a $t$-statistic of $-12.39$ and a $p$-value of $6.43 \times 10^{-12}$, allowing us to reject the null hypothesis that the mean difference between the two score distributions is zero. 

For robustness checks, we also present results for: the partially frozen ResNet-50 model, but with the image-level ROC AUC computed separately for each fold to produce a final fold-averaged score (Supplementary Figure \ref{fig:EF3}); two ResNet-50 models with every hidden layer frozen or unfrozen (Supplementary Figures \ref{fig:EF4}-\ref{fig:EF5}); the partially frozen ResNet-50 model using patient-level predictions instead of image-level predictions (Supplementary Figure \ref{fig:EF6}A); the partially frozen ResNet-50 model run on data in which the minority class (healthy) is oversampled to achieve class balance (the reported outcomes are derived using only 5 random seeds) (Supplementary Figure \ref{fig:EF7}). 

Finally, in the majority of these results, we find that the semantic synthesis models outperform the class-to-image models. This shows that although the LDMs may have similar FID and IS scores, their differentiating factor may be their ability to improve NAFLD classification performance.

\subsubsection*{Performance Improvement over ResNet-50}
A complete NAFLD classification performance evaluation includes testing sensitivity with respect to different classifier architectures as well. Thus, in Figure \ref{fig:figure5}B we compare image-level ROC AUC curves for three families of CNN classifiers: ResNet-50 (our initial architecture), EfficientNet v1, and EfficientNet v2. Similarly to ResNet-50, the two EfficientNet architectures, pretrained on the same ImageNet-1k data, are partially frozen (specifically, the first five convolutional blocks) and trained for 1,000 iterations with a learning rate of $2.0 \times 10^{-5}$ and a batch size of $32$. 
For each model, we show its classification performance when trained on real images only [$r = 0$ (black lines)] and when trained on a mix of semantic synthesis and real images [$r = 1.5$ (red lines)].

Our results indicate that ResNet-50 and EfficientNet v1 perform best when trained on a mix of images (the latter performs slightly better, with an average ROC AUC of 0.884 versus 0.882). We also find that independent of the model architecture, CNNs trained on a mix of images significantly outperform models trained on real images only (paired $t$-test, $p < 0.01$). It should be noted that the choice of random seed, and thus, the choice of synthetic images, has a strong effect on model performance. If we look at the maximum image-level ROC AUC over all random seeds instead of the average, our best EfficientNet v1 model trained on a mix of images has a score of 0.904. This suggests that we could further improve our classification performance by selecting higher quality synthetic images (i.e., by filtering out poor samples such as those in Figure \ref{fig:figure3}) to mix with the real samples. We report patient-level ROC AUC scores (i.e., with predictions averaged across each patient's images) in Supplementary Figure \ref{fig:EF6}B.  Finally, note that we cannot provide a one-to-one comparison with the GAN-based classification results of Che et al., 2021\cite{che2021gan} due to differences in our preprocessing, training, and evaluation methodologies.

\subsubsection*{SHapley Additive exPlanations (SHAP) Analysis}

As mentioned in the Data section, medical professionals rely on key stylistic features in liver ultrasounds to detect and grade NAFLD. In this context, we aim to investigate whether our machine learning framework utilizes these same features to classify NAFLD, with a specific emphasis on the synthesized images. To accomplish this, we employ SHAP, a game-theoretic approach designed to explain the outputs of machine learning models.

In Figure \ref{fig:figure6}, we present four synthetic images (two from the healthy class and two from the unhealthy class) generated using the semantic synthesis LDM, along with their corresponding SHAP values. Within the SHAP heat maps, pixels with positive values are highlighted in red, signifying their positive contributions to NAFLD detection, while pixels with negative values are represented in blue, indicating their negative impact on detection. In the case of the two unhealthy examples (top row in Figure \ref{fig:figure6}), we observe a pronounced concentration of positive SHAP values in the kidney cortex and the body of the liver. This observation suggests that our classifier accurately discerns differences in echogenicity between the liver and the kidney, a feature we discussed in the Data section. While pinpointing features related to blurred veins and diaphragms among the unhealthy examples is challenging, relevant insights can be drawn from the healthy examples. For instance, in the first example from the healthy class (bottom left in Figure \ref{fig:figure6}), the diaphragm is clearly visible and appears to play a substantial role in producing a negative prediction. Furthermore, in the second healthy example (bottom right in Figure \ref{fig:figure6}), negative SHAP values are clustered around the boundaries of multiple portal veins. This implies that well-defined vein and diaphragm boundaries contribute to negative predictions, while their absence indirectly contributes to positive predictions.

\begin{figure}[H]
	\centering
	\includegraphics[scale=0.35]{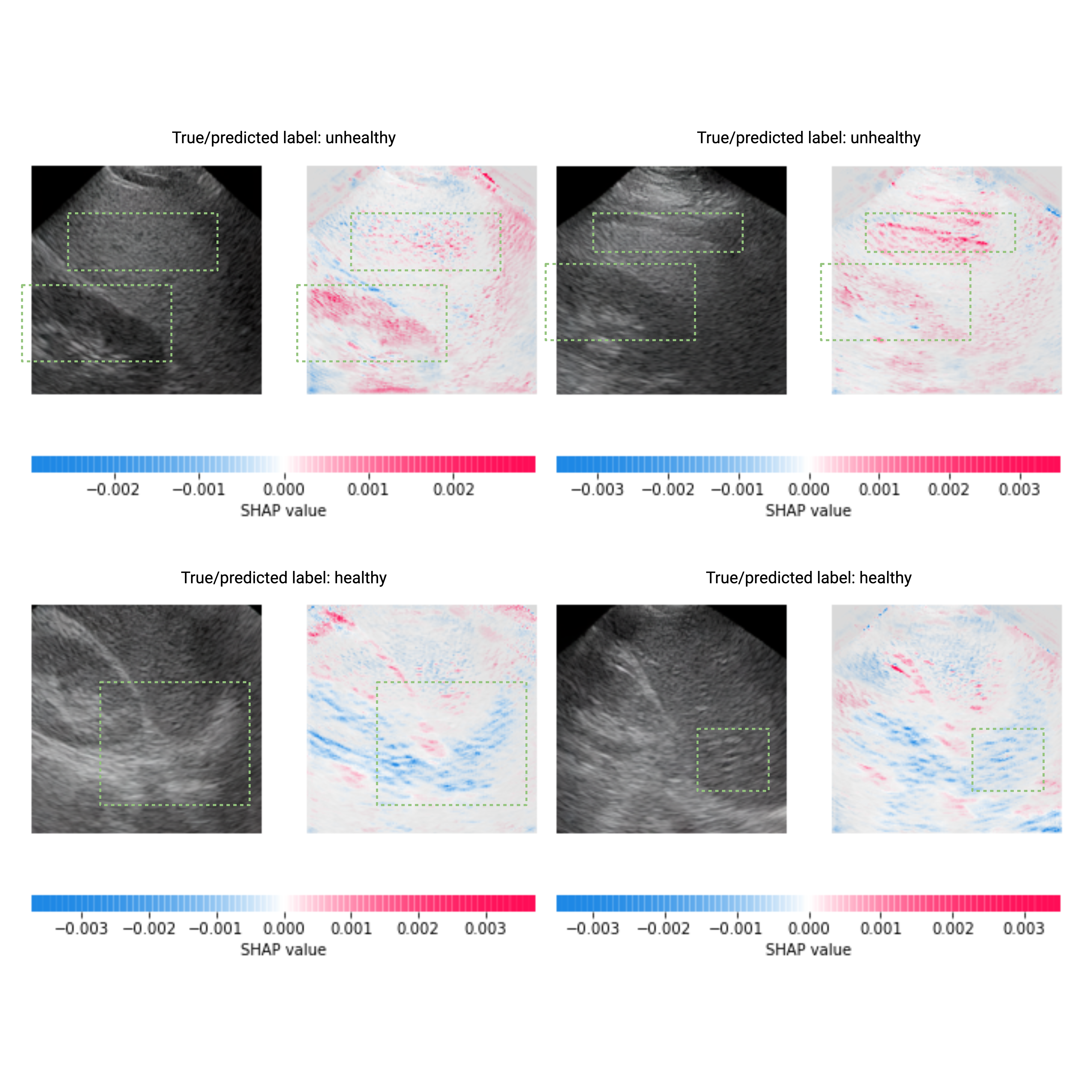}
	\caption{\label{fig:figure6}\textbf{Representative synthesized NAFLD images and their corresponding SHAP values.} The classifier used for this analysis is a CNN trained with a mixing rate of 1.5 using data corresponding to the first seed and cross-validation fold of our machine learning pipeline. All four synthetic examples are generated by the semantic synthesis LDM and are taken from the training set of the classifier. SHAP value regions of particular interest in both the images and the heat maps are highlighted in dotted green rectangles. The top two examples belong to the unhealthy class, while the bottom two examples represent the healthy class.}
 \end{figure}

\section*{\color{Maroon}Discussion}
Our research demonstrates the potential of integrating synthesized images generated by state-of-the-art generative AI algorithms with real liver ultrasound imaging to enhance the detection of non-alcoholic fatty liver disease (NAFLD) in data-constrained environments. By doing so, we aim to contribute toward preventing the progression of this disease to cirrhosis and hepatocellular carcinoma (HCC). Concretely, in this study, we train two latent diffusion models (LDMs) using B-mode liver ultrasound images obtained from 55 patients. One LDM is conditioned on semantic maps, while the other is conditioned on class labels. Our qualitative evaluation of the synthesized images reveals that the LDMs are able to reproduce key stylistic features of NAFLD, and that medical experts are unable to consistently distinguish between real and synthetic liver ultrasound images as demonstrated by the results of an image Turing test. Our experiments also show that the synthetic images generated by our LDMs outperform those generated by generative adversarial networks (GANs) when evaluated using the FID and IS metrics. 

Additionally, we find that combining synthetic and real images significantly enhances the ROC AUC of a CNN backbone classifier fine-tuned for a pixel-level binary NAFLD classification task. This improvement is consistent across both classes of LDM-generated images and multiple CNN architectures. By comparing our approach to traditional geometric augmentation techniques, we determine that LDM-generated images introduce more meaningful variations that facilitate better generalization of classifiers to unseen data.

Although our study showcases advancements over the research conducted by Che et al., 2021\cite{che2021gan}, it is essential to note the limitation of our work, namely the restricted number of images used to train our two diffusion models. To mitigate this issue, we employ random resized crops as a workaround. However, incorporating additional images for training would further enhance the models' capacity to generate diverse, realistic outputs. It may also be beneficial to pretrain the LDMs on other medical imaging datasets before fine-tuning them on a smaller dataset like the one we used. Additionally, it would be valuable to investigate the inclination of LDMs to produce near-exact replicas of training images, as highlighted in the work of Carlini et al., 2023\cite{carlini2023extracting}. This tendency could hinder the models' ability to generate diverse outputs, particularly in low-data regime settings.

Finally, future research should extend this approach to other data modalities to determine whether LDM-generated images can enhance classification performance in broader contexts.

%%%%%%%%%%%%%%%%%%%%%%%%%%
%  BIBLIOGRAPHY
%%%%%%%%%%%%%%%%%%%%%%%%%%
\clearpage
\begin{spacing}{1}
\bibliographystyle{naturemag}
\bibliography{main}

\end{spacing}

%%%%%%%%%%%%%%%%%%%%%%%%%%
%  Author Contributions
%%%%%%%%%%%%%%%%%%%%%%%%%%

\newpage

\bigskip \noindent \textbf{\color{Maroon}Acknowledgements:} We thank Alberto Todeschini and Fred Nugen for their general input on this paper. Our thanks also go to the medical team that participated in our image Turing test. None of the authors have been paid to write this article by a healthcare company or other agency. All authors had full access to the data in the study and accept responsibility to submit for publication.

%We thank Cornelia Ilin for her guidance and supervision over the course of this project. 

\bigskip \noindent \textbf{\color{Maroon}Author contributions:} 
R.H. conceived and led the study. R.H. and R.M. collected and preprocessed the data. J.K. annotated the raw data and confirmed the presence of key stylistic features of NAFLD. R.H. wrote the code to train diffusion models and produce synthetic images. R.H. wrote the classification module to train CNNs on real and synthetic images, with feedback from S.H. C.I. wrote the code to compute SHAP values, which R.H. and J.K. helped interpret for a sample of images. R.H. prepared the contents of the image Turing test, while J.K. recruited participants. R.H. and C.I. wrote the paper, with contributions from J.K. and J.V. on the Introduction and Data sections, respectively. R.H. and C.I designed the figures in the main text and interpreted results. R.H. created the figures shown in the main text and in the appendix. J.K. and C.I. created Supplementary Table \ref{fig:ET0} in Appendix C. R.H., J.K., and C.I. managed literature review. J.K., R.M., S.H., and J.V. all reviewed the final draft and contributed minor edits. R.M., S.H., and J.V. contributed equally and are listed in a randomly assigned order.

\bigskip \noindent \textbf{\color{Maroon}Role of funding source:} The authors declare no funding sources.

\bigskip \noindent \textbf{\color{Maroon}Declaration of interests:} The authors declare no competing interests.

\bigskip \noindent \textbf{\color{Maroon}Tool used to create Figures:} BioRender.com (Figures \ref{fig:figure1}-\ref{fig:figure2})

\bigskip \noindent \textbf{\color{Maroon}Code availability:} The underlying code for this study is not publicly available but may be made available to qualified researchers on reasonable request from the corresponding author.

%%%%%%%%%%%%%%%%%%%%%%%%%%
%  Figures and Tables
%%%%%%%%%%%%%%%%%%%%%%%%%%

\newpage

%%%%%%%%%%%%%%%%%%%%%%%%%%
%  Appendix
%%%%%%%%%%%%%%%%%%%%%%%%%%

\clearpage

\begin{appendices}

\section{Data Summary}
\label{data_summary}

\subsection{Data Analysis}
\label{data_analysis}

\begin{figure}[H]
    %reset figure counter and set name to ST
    \setcounter{figure}{0}
    \renewcommand\figurename{Supplementary Table} 
	\centering
	\includegraphics[scale=0.7]{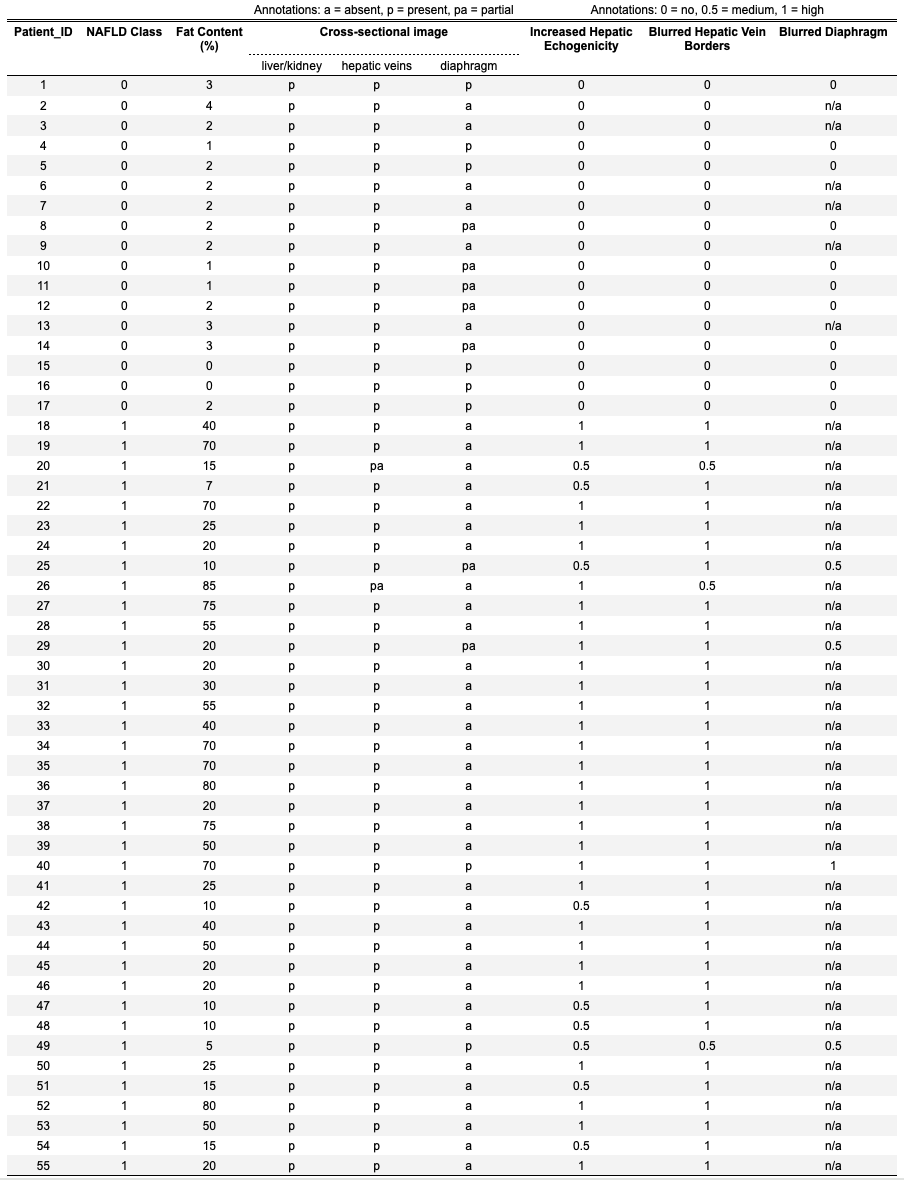}
	\caption{\label{fig:ET0} \textbf{Real ultrasound image analysis.} Columns 1-3 show the ID, disease label, and the percent fat content for each patient, as reported by Byra et al., 2018\cite{byra2018dataset}. Columns 4-6 are binary indicators of the visibility of relevant anatomical structures, annotated by Dr. Joe Klepich. Specifically, these include the liver and kidney, hepatic veins, and diaphragm; ``p" indicates that the structure is present, ``pa" indicates that it is partially visible, and ``a" indicates that it is absent. Columns 7-9 are binary indicators of three stylistic features commonly used to identify patients with NAFLD, also annotated by Dr. Klepich. These features include increased hepatic echogenicity, blurred hepatic vein borders, and a blurred diaphragm: 0 indicates the absence of the feature, 0.5 indicates that the feature is partially or weakly present, and 1 indicates that the feature is strongly present. Whenever an indicator cannot be evaluated for a patient, the corresponding cell is labeled ``n/a."}
\end{figure}

\subsection{Data Preprocessing}
\label{data_preprocessing}
Although the preprocessing transformations are mostly the same for the diffusion and classification machine learning pipelines, there are slight differences between the two approaches. In the diffusion pipeline, image preprocessing is applied dynamically during training, thus if a raw image appears in $n$ mini-batches it will generate $n$ cropped and resized images. In the classification pipeline, the preprocessing is fixed beforehand, such that each raw image corresponds to a single preprocessed image. The preprocessed dataset is saved and reused across all classification experiments so that we can objectively evaluate the impact of synthetic images on model performance. Also, in keeping with the pretraining schemes of our CNN classifier backbones, each image is normalized using the mean and standard deviation of the ImageNet-1k dataset (Deng et al., 2009\cite{deng2009imagenet}).

\section{Methods Summary}
\label{methods_summary}

\subsection{Latent Diffusion Models}
\label{latent_diffusion_models}
Latent diffusion models (LDMs) are variants of diffusion models that first encode images to a low-dimensional latent space representation using a pretrained encoder $E$, thus avoiding costly computations in pixel space. A visual diagram of a LDM is shown in Rombarch et al., 2021\cite{rombach2021ldm}. During training, an image $x$ is fed through $E$ to produce a latent vector $z$. The model then adds Gaussian noise to $z$ according to a fixed Markov chain, producing $z_T$. The training objective of the model is to learn the reverse process, that is, how to denoise $z_T$ back to its original state $z$. Mathematically, this objective can be expressed as:
\begin{equation}
L_{\text{LDM}} := \mathbb{E}_{E(x),\epsilon\sim\mathcal{N}(0, 1),t}\bigg[\|\epsilon - \epsilon_\theta(z_t, t, \tau_\theta(y))\|_2^2\bigg], 
\end{equation}
where $\epsilon_\theta$ is a denoising autoencoder (shown in the figure as a UNet model) and $\tau_\theta$ is an encoder that feeds conditioning inputs $y$ (i.e. text, semantic maps, and class labels) to $\epsilon_\theta$ via concatenation and cross-attention layers. Finally, the recovered latent vector $\tilde{z}$ can be projected back to pixel space through the use of a pretrained decoder $D$.

\section{Supplementary Figures}
% reset figure coutner and figure name
\setcounter{figure}{0}
\renewcommand\figurename{Supplementary Figure} 

\begin{figure}[H]
	\centering
	\includegraphics[scale=0.75]{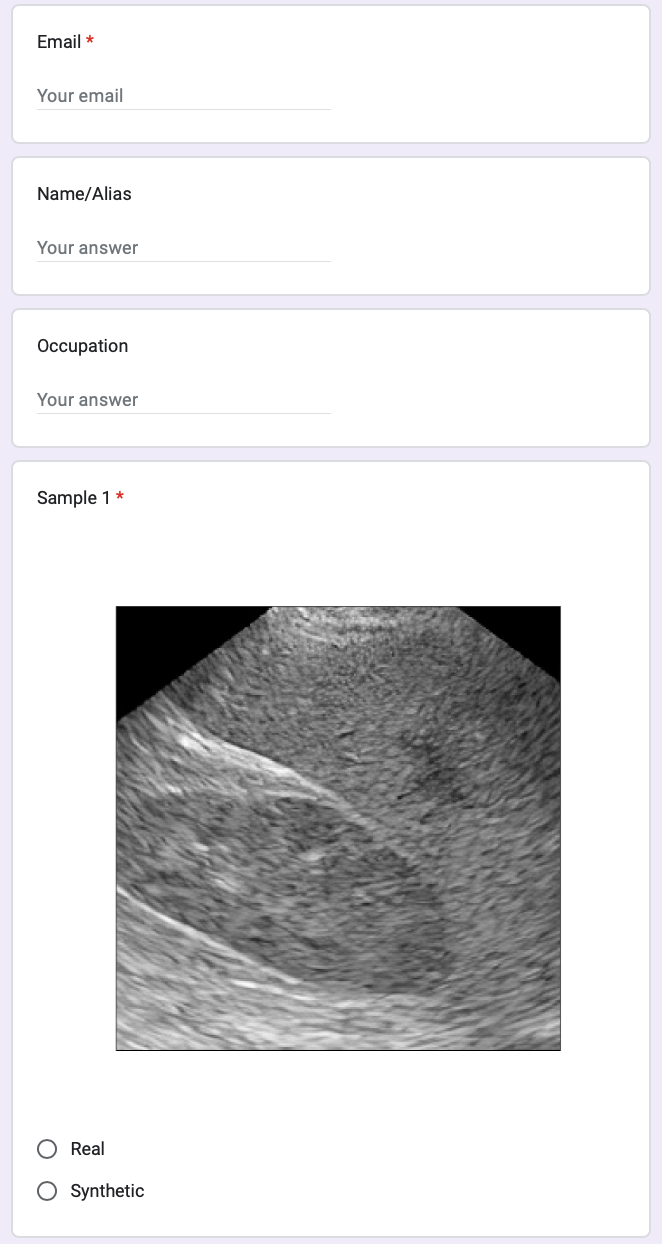}
	\caption{\label{fig:EF1} \textbf{Turing test webpage.} A panel comprising five medical experts evaluated the realism (``Real" vs. ``Synthetic") of 50 randomly selected images from our real and synthesized ultrasound databases. Image Sample 1 is shown here as an example.}
\end{figure}

\begin{figure}[H]
	\centering
	\includegraphics[scale=0.25]{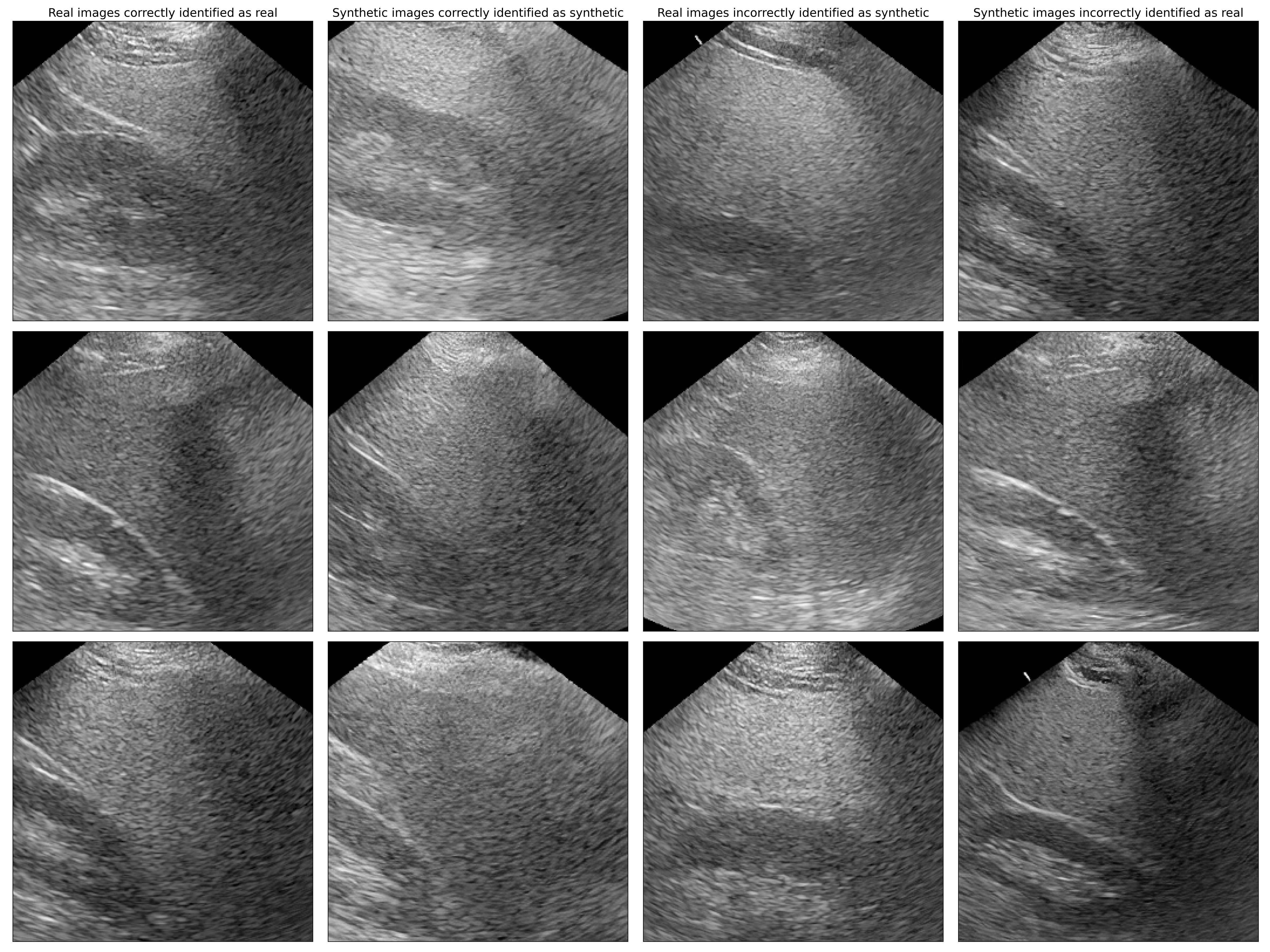}
	\caption{\label{fig:EF2} \textbf{Top correct and incorrect samples from the Turing test.} The first two columns show the top three real and synthetic images most often correctly identified by test participants. The third and fourth columns show the top three real and synthetic images most often incorrectly identified by test participants.}
\end{figure}

\begin{figure}[H]
	\centering
	\includegraphics[scale=0.35]{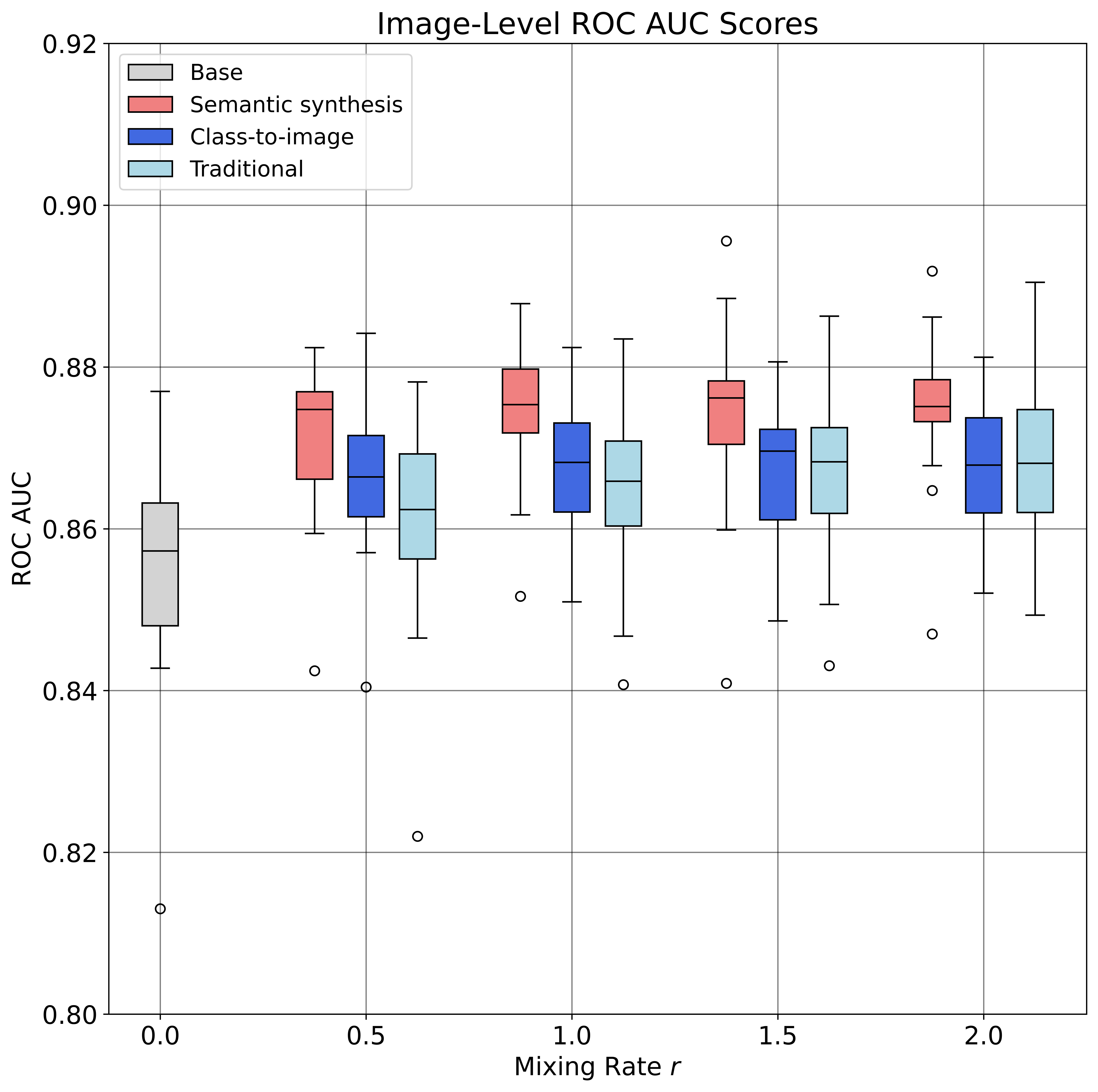}
	\caption{\label{fig:EF3} \textbf{Image-level out-of-sample NAFLD classification performance -- Sensitivity test 1.} These results are generated using the same ResNet-50 model as in Figure \ref{fig:figure5}A. However, instead of concatenating the out-of-fold predictions and computing the ROC AUC, we average the ROC AUC across the five folds to produce a final fold-averaged score.}
\end{figure}

\begin{figure}[H]
	\centering
	\includegraphics[scale=0.35]{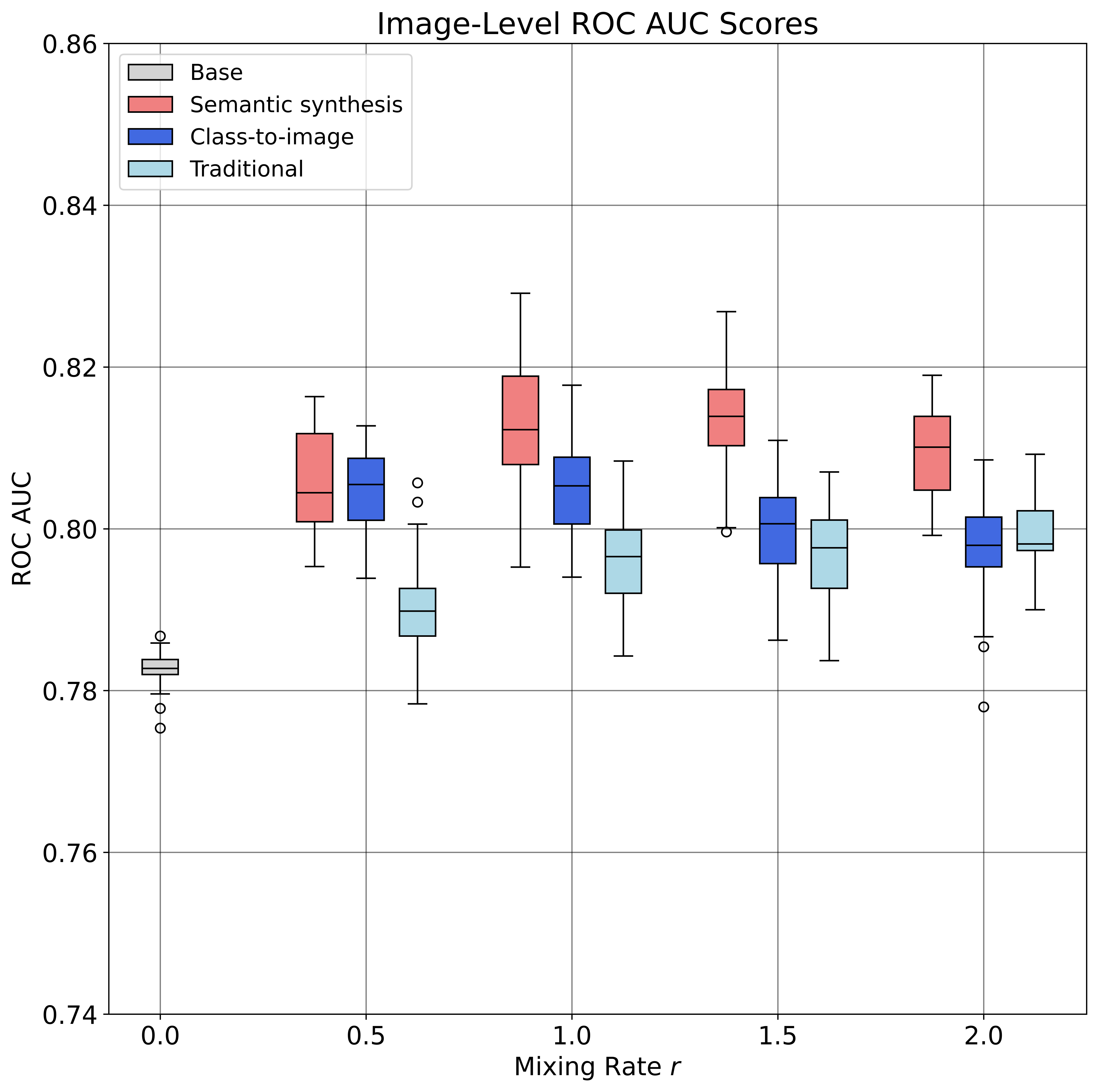}
	\caption{\label{fig:EF4} \textbf{Image-level out-of-sample NAFLD classification performance -- Sensitivity test 2.} This plot shows results for the same data input scenarios as in Figure \ref{fig:figure5}A, except that we freeze every hidden layer in the ResNet-50 backbone.}
\end{figure}

\begin{figure}[H]
	\centering
	\includegraphics[scale=0.35]{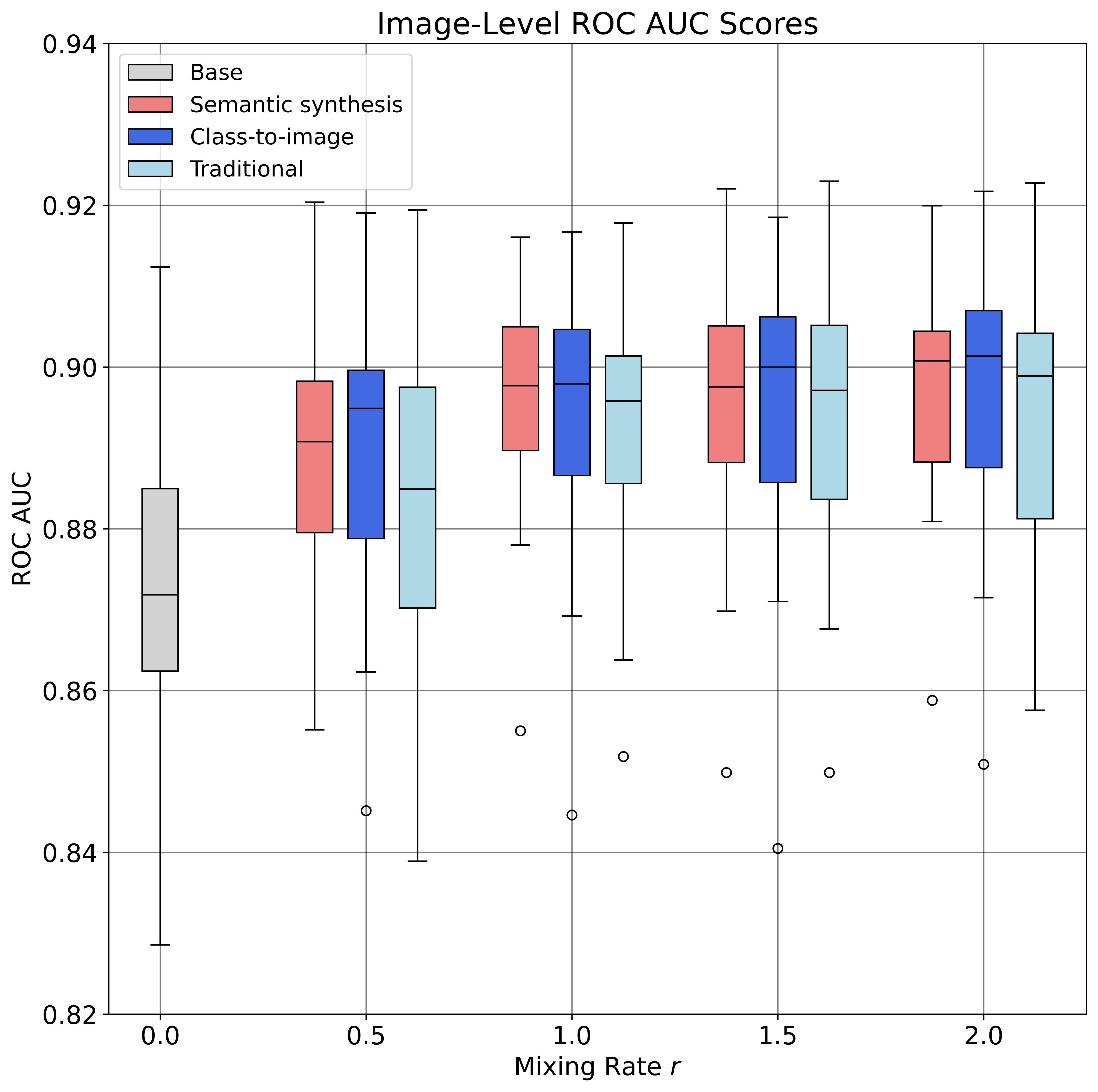}
	\caption{\label{fig:EF5}\textbf{Image-level out-of-sample NAFLD classification performance -- Sensitivity test 3.} This plot shows results for the same data input scenarios as in Figure \ref{fig:figure5}A, except that we unfreeze every hidden layer in the ResNet-50 backbone.}
\end{figure}

\begin{figure}[H]
	\centering
	\includegraphics[scale=0.35]{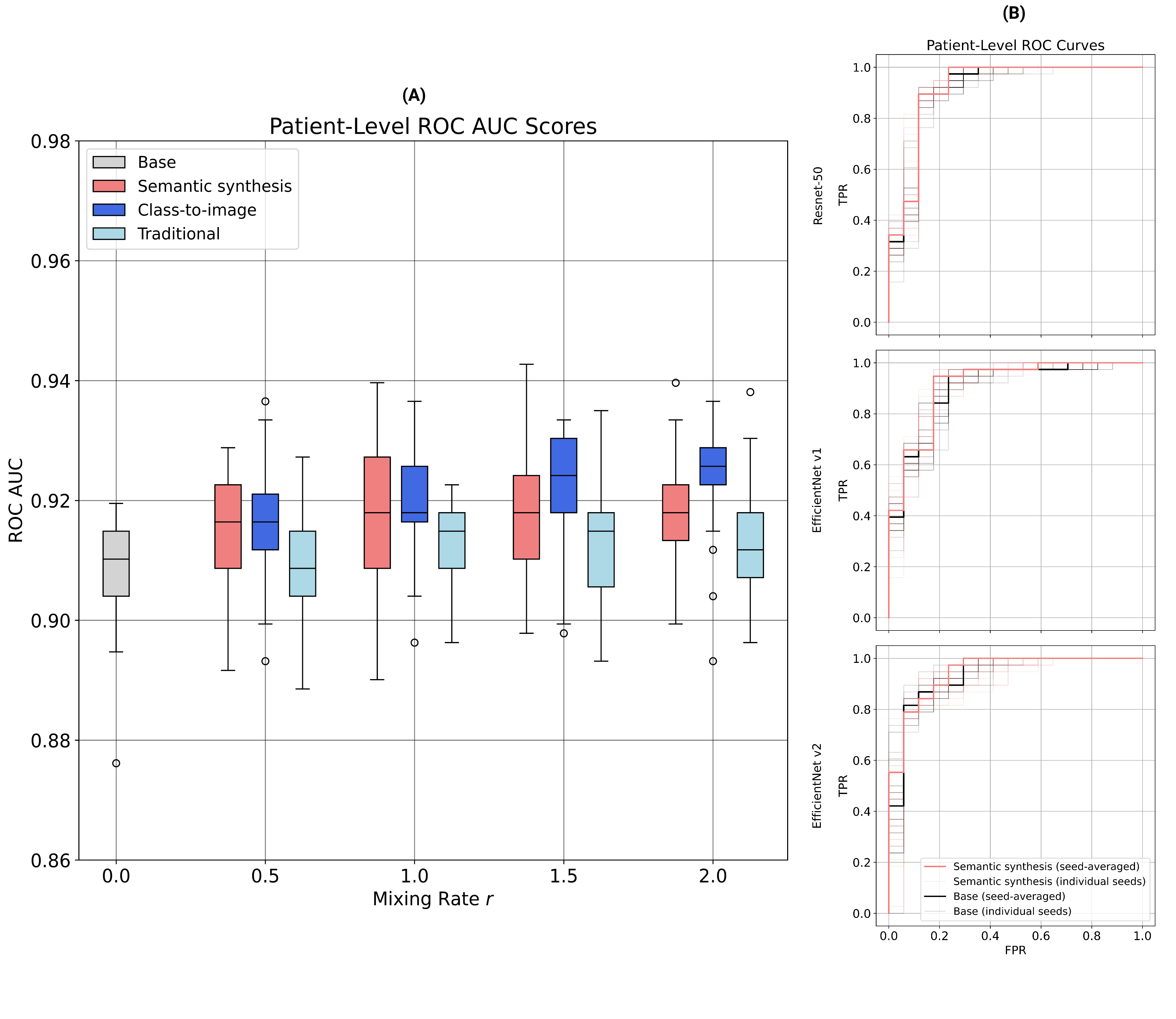}
	\caption{\label{fig:EF6}\textbf{Patient-level out-of-sample NAFLD classification performance -- Sensitivity test 4}. \textbf{(A)} Box plots of the patient-level ROC AUC as a function of the mixing rate $r$. \textbf{(B)} Patient-level ROC curves for three families of CNN classifiers: ResNet-50 (top), EfficientNet v1 (middle), and EfficientNet v2 (bottom). The average ROC AUC for the base and synthetically augmented models are as follows: ResNet-50, 0.907 versus 0.918; EfficientNet v1, 0.896 versus 0.908; EfficientNet v2, 0.932 versus 0.927.}
\end{figure}

\begin{figure}[H]
	\centering
	\includegraphics[scale=0.35]{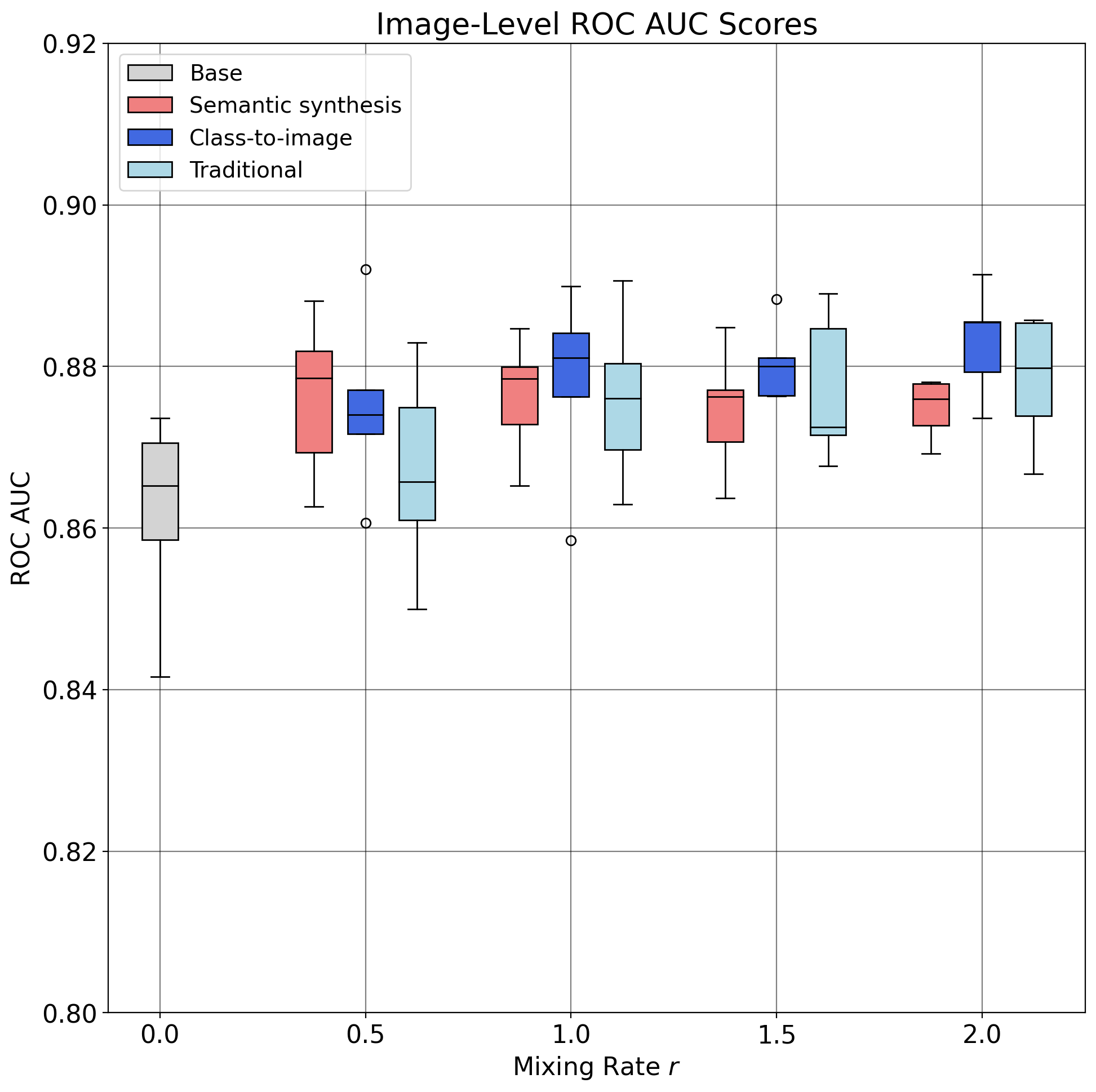}
	\caption{\label{fig:EF7}\textbf{Image-level out-of-sample NAFLD classification performance -- Sensitivity test 5}. Image-level ROC AUC as a function of the mixing rate $r$ for five random seeds using balanced classes. Data balancing is done by upsampling the minority (healthy) class.}
\end{figure}

\end{appendices}

\end{document}